\documentclass[12pt]{article}
\usepackage{graphicx}
\usepackage{verbatim}
\usepackage{amssymb}
\usepackage{amsbsy}
\usepackage{amscd}
\usepackage{amsmath}
\usepackage{amsthm}
\newtheorem{Theorem}{{\bf Theorem}}
\newtheorem{Lemma}{{\bf Lemma}}
\newtheorem{Corollary}{{\bf Corollary}}

\newcommand{\RR}{\mathbb R}
\newcommand{\CC}{\mathbb C}
\newcommand{\EE}{\mathbb E}
\newenvironment{keywords}{{\bf Keywords:} }{\\}

\title{Asymptotic Equivalence of Bayes Cross Validation \\and Widely Applicable Information 
Criterion \\ in Singular Learning Theory}
\author{Sumio Watanabe\\
P\&I Lab., Tokyo Institute of Technology\\
4259 Nagatsuta, Midoriku, Yokohama, 226-8503 Japan\\
E-mail: swatanab(at)pi.titech.ac.jp\\
P\&I Lab., Tokyo Institute of Technology\\
4259 Nagatsuta, Midoriku, Yokohama, 226-8503 Japan}

\begin{document}

\maketitle

\begin{abstract}
In regular statistical models, 
the leave-one-out cross-validation is asymptotically equivalent to 
the Akaike information criterion. However, since many learning machines 
are singular statistical models, 
the asymptotic behavior of the cross-validation remains unknown.
In previous studies, we established the singular learning theory and 
proposed a widely applicable information criterion, the 
expectation value of which is asymptotically equal to the average Bayes generalization loss. 
In the present paper, we theoretically compare the Bayes cross-validation loss and the widely 
applicable information criterion and prove two theorems. 
First, the Bayes cross-validation loss is asymptotically equivalent to the 
widely applicable information criterion as a random variable. 
Therefore, model selection and hyperparameter optimization using these two values 
are asymptotically equivalent. 
Second, the sum of 
the Bayes generalization error and the Bayes cross-validation error 
is asymptotically equal to $2\lambda/n$, where $\lambda$ 
is the real log canonical threshold and $n$ is the number of
training samples. Therefore the relation between 
the cross-validation error and the generalization error is determined by the
algebraic geometrical structure of a learning machine. 
We also clarify that the deviance information criteria are different from 
the Bayes cross-validation and the widely applicable information criterion. 
\end{abstract}


\section{Introduction}

A statistical model or a learning machine is said to be 
regular if the map taking parameters to probability
distributions is one-to-one and if its Fisher information
matrix is positive definite. 
If a model is not regular, then it is said to be singular. 
Many learning machines, such as 
artificial neural networks \cite{NN2001}, 
normal mixtures \cite{Yamazaki}, 
reduced rank regressions \cite{Aoyagi}, 
Bayes networks \cite{Rusakov,Zwiernik}, 
mixtures of probability distributions \cite{Lin}, 
Boltzmann machines \cite{Aoyagi2}, and 
hidden Markov models \cite{Yamazaki2}, are not regular but singular \cite{IEEE2007}. 
If a statistical model or a learning machine
contains a hierarchical structure, hidden variables, or a grammatical
rule, then the model is generally singular. Therefore, singular learning theory 
is necessary in modern information science. 

The statistical properties of singular models
have remained unknown until recently,
because analyzing a singular likelihood function 
had been difficult \cite{Hartigan, NOLTA1995}. In singular statistical
models, the maximum likelihood estimator does not satisfy
asymptotic normality. Consequently, AIC 
is not equal to the average generalization error \cite{Hagiwara}, and 
the Bayes information criterion (BIC) is not equal to the 
Bayes marginal likelihood \cite{NC2001}, even asymptotically. 
In singular models, the maximum likelihood estimator often diverges, or
even if it does not diverge, makes the generalization error very large. 
Therefore, the maximum likelihood method is not appropriate for singular models. 
On the other hand, Bayes estimation was proven to make the generalization 
error smaller if the statistical model contains singularities. Therefore, in the present paper, 
we investigate methods for estimating the Bayes generalization error.

Recently, new statistical learning theory, 
based on methods from algebraic geometry, has been established 
\cite{NC2001, Drton, Cambridge2009, NN2010, ASPM2010, Lin}.
In singular learning theory, a log likelihood function 
can be made into a common standard form, even if it contains 
singularities, by using the resolution theorem in algebraic geometry. 
As a result, the asymptotic behavior of the posterior distribution is 
clarified, and the concepts of BIC and AIC can be generalized 
onto singular statistical models. The asymptotic Bayes marginal likelihood was proven to 
be determined by the real log canonical threshold \cite{NC2001}, and 
the average Bayes generalization error was proven to be estimable by the
widely applicable information criterion \cite{Cambridge2009, NN2010, ASPM2010}.

Cross-validation is an alternative method for estimating
the generalization error \cite{Mosier, Stone, Geisser}. 
By definition, the average of the cross-validation is 
equal to the average generalization error in both regular
and singular models. In regular statistical models, the leave-one-out 
cross-validation is asymptotically equivalent to AIC \cite{Akaike} 
in the maximum likelihood method \cite{Stone, Linhart, Browne}. 
However, the asymptotic behavior of the cross-validation in singular models has not been clarified. 

In the present paper, in singular statistical models, we theoretically 
compare the Bayes cross-validation, the widely applicable information
criterion, and the Bayes generalization error and prove two theorems.
First, we show that the Bayes cross-validation loss is asymptotically 
equivalent to the widely applicable information criterion as a random variable. 
Second, we also show that the sum of the 
Bayes cross-validation error and the Bayes generalization error is 
asymptotically equal to $2\lambda/n$, where $\lambda$ is 
the real log canonical threshold and $n$ is the number of training samples. 
It is important that neither $\lambda$ or $n$ is a random variable. 
Since the real log canonical threshold is 
a birational invariant of the statistical model, the relationship between the Bayes
cross-validation and the Bayes generalization error is determined by 
the algebraic geometrical structure of the statistical model. 

The remainder of the present paper is organized as follows. In Section 2, 
we introduce the framework of Bayes learning and explain 
singular learning theory. In Section 3, 
the Bayes cross-validation is defined. In Section 4,
the main theorems are proven. In Section 5, we discuss
the results of the present paper, and the differences among the cross-validation, 
the widely applicable information criterion, and the deviance 
information criterion are investigated theoretically and experimentally. 
Finally, in Section 6, we summarize the primary conclusions of the present paper.

\begin{table}[tb]\label{table:111}
\begin{center}
\begin{tabular}{|c|c|c|}
\hline
Variable & Name & eq.  number \\
\hline
$\EE_{w}[\;\;]$ & posterior average &eq.(\ref{eq:BayesP}) \\
$\EE_{w}^{(i)}[\;\;]$ & posterior average without $X_{i}$ &
eq.(\ref{eq:Ew(i)}) \\
\hline 
$L(w)$ & log loss function &eq.(\ref{eq:L(w)})\\
$L_{0}$ & minimum loss & eq.(\ref{eq:L0}) \\
$L_{n}$ & empirical loss & eq.(\ref{eq:Ln}) \\
\hline
$B_{g}L(n)$ & Bayes generalization loss &eq.(\ref{eq:Blg})\\
$B_{t}L(n)$ & Bayes training loss &eq.(\ref{eq:Blt}) \\
$G_{t}L(n)$ & Gibbs training loss & eq.(\ref{eq:Gibbs})\\
$C_{v}L(n)$ & cross-validation loss &eq.(\ref{eq:Clv}) \\
\hline
$B_{g}(n)$ & Bayes generalization error &eq.(\ref{eq:Bg})\\
$B_{t}(n)$ & Bayes training error &eq.(\ref{eq:Bt}) \\
$C_{v}(n)$ & cross-validation error & eq.(\ref{eq:Cv}) \\
\hline
$V(n)$ & functional variance & eq.(\ref{eq:V(n)}) \\
$Y_{k}(n)$ & $k$th functional cumulant & eq.(\ref{eq:Yk}) \\
\hline
$\mbox{WAIC}(n)$ & WAIC &eq.(\ref{eq:WAIC})\\
\hline
$\lambda$ & real log canonical threshold &eq.(\ref{eq:lambda})\\
$\nu$ & singular fluctuation &eq.(\ref{eq:nyu})\\
\hline
\end{tabular}
\end{center}
\caption{Variables, Names, and Equation Numbers}
\end{table}

\section{Bayes Learning Theory} 

In this section, we summarize Bayes learning theory for singular learning machines. 
The results presented in this section are well known and are the fundamental basis of
the present  paper. Table \ref{table:111} lists variables, names, and
equation numbers in the present paper. 

\subsection{Framework of Bayes Learning}

First, we explain the framework of Bayes learning. 

Let $q(x)$ be a probability density function on
the $N$ dimensional real Euclidean space ${\RR}^{N}$. 
The training samples and the testing sample
are denoted by random variables 
$X_{1},X_{2},...,X_{n}$ and $X$, respectively, which
are independently subject to the
same probability distribution as $q(x)dx$. 
The probability distribution $q(x)dx$ is sometimes called the true 
distribution. 

A statistical model or a learning machine is defined as 
a probability density function $p(x|w)$ of $x\in {\RR}^{N}$ for a given 
parameter $w\in W\subset {\RR}^{d}$, where $W$ is the set of all parameters. 
In Bayes estimation, we prepare a probability density function $\varphi(w)$ on
$W$. Although $\varphi(w)$ is referred to as a prior distribution, in general, $\varphi(w)$
does not necessary represent an {\it a priori} knowledge of 
the parameter. 

For a given function $f(w)$ on $W$, the expectation value 
of $f(w)$ with respect to the posterior distribution is defined as
\begin{equation}\label{eq:BayesP}
\EE_{w}[f(w)] = \frac{\displaystyle
\int f(w)\;\prod_{i=1}^{n}p(X_{i}|w)^{\beta}\;\varphi(w)dw
}{\displaystyle
\int \prod_{i=1}^{n}p(X_{i}|w)^{\beta}\;\varphi(w)dw
},
\end{equation} 
where $0<\beta<\infty$ is the inverse temperature. 
The case in which $\beta=1$ is most important because this case 
corresponds to strict Bayes estimation. 
The Bayes predictive distribution is defined as 
\begin{equation}\label{eq:predictive}
p^{*}(x)\equiv \EE_{w}[ p(x|w)]. 
\end{equation}
In Bayes learning theory, the following random variables 
are important. The Bayes generalization loss $B_{g}L(n)$ and 
the Bayes training loss $B_{t}L(n)$ are 
defined, respectively, as
\begin{eqnarray}\label{eq:Blg}
B_{g}L(n)&=&-\EE_{X}[\log  p^{*}(X) ],\\
B_{t}L(n)&=&-\frac{1}{n}\sum_{i=1}^{n}\log p^{*}(X_{i}),\label{eq:Blt}
\end{eqnarray}
where $\EE_{X}[\;\;]$ gives the expectation value over $X$. 
The {\it functional variance} is defined as 
\begin{equation}\label{eq:V(n)}
V(n)=\sum_{i=1}^{n}\Bigl\{
\EE_{w}[(\log p(X_{i}|w))^{2}]
-\EE_{w}[\log p(X_{i}|w)]^{2}\Bigr\},
\end{equation}
which shows the fluctuation of the posterior distribution.
In previous papers \cite{Cambridge2009,NN2010,IEICE2010},
we defined the widely applicable information criterion
\begin{equation}\label{eq:WAIC}
\mbox{WAIC}(n)\equiv B_{t}L(n)+\frac{\beta}{n}V(n),
\end{equation}
and proved that 
\begin{equation}\label{eq:eq-of-state}
\EE[B_{g}L(n)]=\EE[\mbox{WAIC}(n)]+o(\frac{1}{n}),
\end{equation}
holds for both regular and singular statistical models, 
where $\EE[\;\;]$ gives the expectation value over the sets of
training samples. 
\vskip5mm\noindent
{\bf Remark.} Although the case in which $\beta=1$ is most important, 
general cases in which $0<\beta<\infty$ are also important for four reasons. 
First, from a theoretical viewpoint, 
several mathematical relations can be obtained using the derivative of $\beta$. 
For example, using the Bayes free energy or 
the Bayes stochastic complexity,
\begin{equation}
{\cal F}(\beta)=-\log\int \prod_{i=1}^{n}p(X_{i}|w)^{\beta}\varphi(w)dw,
\end{equation}
the Gibbs training loss
\begin{equation}\label{eq:Gibbs}
G_{t}L(n)=-\EE_{w}\Bigl{[}
\frac{1}{n}\sum_{i=1}^{n}\log p(X_{i}|w)
\Bigr{]}
\end{equation}
can be written as
\begin{equation}
G_{t}L(n)=\frac{\partial {\cal F}}{\partial \beta}.
\end{equation}
Such relations are useful in investigating Bayes learning theory. 
We use $\partial^{2}{\cal F}/\partial \beta^{2}$ to investigate 
the deviance information criteria
in Section \ref{section:discuss}.
Second, the maximum likelihood method formally corresponds 
to $\beta=\infty$. The maximum likelihood method is 
defined as 
\begin{equation}
p^{*}(x)=p(x|\hat{w}),
\end{equation}
instead of eq. (\ref{eq:predictive}), 
where $\hat{w}$ is the maximum likelihood estimator. 
Its generalization loss is also defined in the same manner as eq. (\ref{eq:Blg}). 
In regular statistical models, the asymptotic 
Bayes generalization error does not depend on $0<\beta\leq \infty$,
whereas in singular models it strongly depends on $\beta$. 
Therefore, the general case is useful for investigating the difference between 
the maximum likelihood and Bayes methods. 
Third, from an experimental viewpoint, 
in order to approximate the posterior distribution, 
the Markov chain Monte Carlo method is often 
applied by controlling $\beta$. In particular, the identity 
\begin{equation}
{\cal F}(1)=\int_{0}^{1}\frac{\partial F}{\partial \beta}\;d\beta
\end{equation}
is used in the calculation of the Bayes marginal likelihood.
The theoretical results for general $\beta$ are useful for monitoring the effect 
of controlling $\beta$ \cite{Nagata}. 
Finally, in the regression problem, $\beta$ can be understood 
as the variance of the unknown additional noise \cite{ASPM2010}
and so may be optimized 
as the hyperparameter. For these reasons, in the present paper, we theoretically 
investigate the cases for general $\beta$. 

\subsection{Notation}

In the following, we explain the notation used in the present study. 

The log loss function $L(w)$ and the entropy $S$ of the true distribution
are defined, respectively, as
\begin{eqnarray}
L(w)&=&-\EE_{X}[\log p(X|w)],\label{eq:L(w)}\\
S&=&-\EE_{X}[\log q(X)].
\end{eqnarray}
Then, $L(w)=S+D(q||p_{w})$, where $D(q||p_{w})$ is 
the Kullback-Leibler distance defined as
\begin{equation}
D(q||p_{w})=\int q(x)\log \frac{q(x)}{p(x|w)}dx.
\end{equation}
Then, $D(q||p_{w})\geq 0$, hence $L(w)\geq S$. Moreover, $L(w)=S$ 
if and only if $p(x|w)=q(x)$. 

In the present paper, 
we assume that there exists a parameter $w_{0}\in W$ 
that minimizes $L(w)$, 
\begin{equation}
L(w_{0})=
\min_{w\in W} L(w).
\end{equation}
Note that such $w_{0}$ is not unique in general because the map 
$w\mapsto p(x|w)$ is, in general, not a one-to-one map in singular learning machines. 
In addition, we assume that, for an arbitrary $w$ that satisfies $L(w)=L(w_{0})$, $p(x|w)$ is 
the same probability density function. Let $p_{0}(x)$ be such a unique probability 
density function. In general, the set
\begin{equation}
W_{0}=\{w\in W; p(x|w)=p_{0}(x)\}
\end{equation}
is not a set of a single element but rather
an analytic or algebraic set with singularities.
Here, 
a set in $\RR^{d}$ is said to be an analytic or algebraic set if and only if the set is equal to the
set of all zero points of an analytic or algebraic function, respectively.
For simple notations, the minimum log loss $L_{0}$ and the empirical log loss $L_{n}$ are
defined, respectively, as
\begin{eqnarray}
L_{0}&=&-\EE_{X}[\log p_{0}(X)], \label{eq:L0}\\
L_{n}&=&-\frac{1}{n}\sum_{i=1}^{n}\log p_{0}(X_{i}).\label{eq:Ln}
\end{eqnarray}
Then, by definition, $L_{0}=\EE[L_{n}]$. 
Using these values, Bayes generalization error $B_{g}(n)$ and 
Bayes training error $B_{t}(n)$ are defined, respectively, as
\begin{eqnarray}
B_{g}(n)&=&B_{g}L(n)-L_{0},\label{eq:Bg}\\ 
B_{t}(n)&=&B_{t}L(n)-L_{n}.\label{eq:Bt}
\end{eqnarray}
Let us define a log density ratio function as:
\begin{equation}
f(x,w)=\log\frac{p_{0}(x)}{p(x|w)},
\end{equation}
which is equivalent to 
\begin{equation}
p(x|w)=p_{0}(x)\exp(-f(x,w)).
\end{equation}
Then, it immediately follows that 
\begin{eqnarray}
B_{g}(n)&=&-\EE_{X}[\log \EE_{w}[ \exp(-f(X,w))] ],\\
B_{t}(n)&=&-\frac{1}{n}\sum_{i=1}^{n}\log \EE_{w}[ \exp(-f(X_{i},w))],\\
V(n)&=& \sum_{i=1}^{n}\Bigl\{
\EE_{w}[f(X_{i},w)^{2}]
-\EE_{w}[f(X_{i},w)]^{2}\Bigr\}.
\end{eqnarray}
Therefore, the problem of statistical learning is characterized by the function $f(x,w)$. 
\vskip5mm\noindent
{\bf Definition}. \\
(1) If 
$q(x)=p_{0}(x)$, then $q(x)$ is said to be {\it realizable} by $p(x|w)$.
Otherwise, $q(x)$ is said to be {\it unrealizable}.\\
(2) If the set $W_{0}$ consists of a single point $w_{0}$ and if 
the Hessian matrix $\nabla\nabla L(w_{0})$ is strictly positive definite, then 
$q(x)$ is said to be {\it regular} for $p(x|w)$. 
Otherwise, $q(x)$ is said to be {\it singular} for $p(x|w)$. 
\vskip5mm\noindent
Bayes learning theory was investigated for a realizable and regular case
\cite{Schwarz,Levin,Amari}. The WAIC was found 
for a realizable and singular case
\cite{NC2001,Cambridge2009,NN2010} and for an 
unrealizable and regular case \cite{IEICE2010}. 
In addition, WAIC was generalized for an unrealizable and singular case \cite{IWSMI2010}.

\subsection{Singular Learning Theory}\label{subsection:assumption}

We summarize singular learning theory. In the present paper, 
we assume the followings. 
\vskip5mm\noindent
{\bf Assumptions.} \\
(1) The set of parameters $W$ is a compact set in ${\RR}^{d}$, the open
kernel \footnote{The open kernel of a set $A$ is the
largest open set that is contained in $A$.} of which is not the empty set. 
The boundary of $W$ is defined by several analytic functions, 
\begin{equation}
W=\{w\in {\RR}^{d};\pi_{1}(w)\geq 0,\pi_{2}(w)\geq 0,...,
\pi_{k}(w)\geq 0\}.
\end{equation}
(2) The prior distribution satisfies $\varphi(w)=\varphi_{1}(w)\varphi_{2}(w)$, 
where $\varphi_{1}(w)\geq 0$ is an analytic function and $\varphi_{2}(w)>0$ is
a $C^{\infty}$-class function. \\
(3) Let $s\geq 8$ and let
\begin{equation}
L^{s}(q)=\{f(x);\|f\|\equiv \Bigl(\int |f(x)|^{s}q(x)dx\Bigr)^{1/s}<\infty\}
\end{equation}
be a Banach space. The map $W\ni w\mapsto f(x,w)$ is 
an $L^{s}(q)$ valued analytic function. \\
(4) A nonnegative function $K(w)$ is defined as 
\begin{equation}
K(w)=\EE_{X}[f(X,w)].
\end{equation}
The set $W_{\epsilon}$ is defined as
\begin{equation}
W_{\epsilon}=\{w\in W\;;\;K(w)\leq \epsilon\}.
\end{equation}
It is assumed that there exist constants $\epsilon,c>0$ such that
\begin{equation}\label{eq:finite}
(\forall w\in W_{\epsilon})\;\;\;
\EE_{X}[f(X,w)]\geq c\;\EE_{X}[f(X,w)^{2}].
\end{equation}
\vskip3mm\noindent
{\bf Remark.} In ordinary learning problems, 
if the true distribution is regular for
or realizable by a learning machine, then assumptions (1), (2), (3) and (4) are satisfied,
and the results of the present paper hold. If the true distribution
is singular for and unrealizable by a learning machine, then
assumption (4) is satisfied in some cases but not in other cases. 
If the assumption (4) is not satisfied, then the Bayes generalization and 
training errors may have asymptotic behaviors other than those described in 
Lemma \ref{Lemma:1} \cite{IWSMI2010}. 
\vskip10mm
The investigation of cross-validation in singular learning machines requires 
singular learning theory. In previous papers,
we obtained the following lemma. 
\begin{Lemma}\label{Lemma:1}
Assume that assumptions (1), (2), (3), and (4) are satisfied. Then, the 
followings hold.\\ 
(1) Three random variables $nB_{g}(n)$, $nB_{t}(n)$,
and $V(n)$ converge in law, when $n$ tends to infinity. 
In addition, the expectation values of these variables converge. 
\\
(2) For $k=1,2,3,4$, we define 
\begin{equation}
M_{k}(n)\equiv \sup_{|\alpha|\leq 1+\beta }
\EE \Bigl{[}\frac{1}{n}\sum_{i=1}^{n}
\frac{\EE_{w}[|f(X_{i},w)|^{k}\exp(\alpha f(X_{i},w))]}
{\EE_{w}[\exp(\alpha f(X_{i},w))]}
\;\Bigr{]},
\end{equation}
where $\EE[\;\;]$ gives the average over all sets of training samples.
Then, 
\begin{equation}\label{eq:assume}
{\rm limsup}_{n\rightarrow\infty }\Bigl(n^{k/2}\;M_{k}(n)\Bigr)<\infty.
\end{equation}
(3) The expectation value of the Bayes generalization loss is asymptotically equal to 
the widely applicable information criterion, 
\begin{equation}\label{eq:BLWAIC}
\EE[B_{g}L(n)]=\EE[\mbox{WAIC}(n)]+o(\frac{1}{n}). 
\end{equation}
\end{Lemma}
\noindent(Proof) For the case in which $q(x)$ is realizable by and singular for
$p(x|w)$, this lemma was proven in \cite{NN2010,Cambridge2009}. In fact, 
the proof of Lemma \ref{Lemma:1} (1) is given in Theorem 1 of \cite{NN2010}. 
Also Lemma \ref{Lemma:1} (2) can be proven in the same manner as 
eq. (32) in \cite{NN2010} or eq. (6.59) in \cite{Cambridge2009}.
The proof of Lemma \ref{Lemma:1} (3) is given in Theorem 2 and 
the discussion of \cite{NN2010}. 
For the case in which $q(x)$ is regular for and unrealizable by $p(x|w)$, 
this lemma was proven in \cite{IEICE2010}. For the case in which 
$q(x)$ is singular for and unrealizable by $p(x|w)$, 
these results can be generalized under the condition that
eq.(\ref{eq:finite}) is satisfied \cite{IWSMI2010}. 
(Q.E.D.)

\section{Bayes Cross-validation}

In this section, we introduce the cross-validation in Bayes learning. 

The expectation value $\EE_{w}^{(i)}[\;\;\;]$ 
using the 
posterior distribution leaving out $X_{i}$ is 
defined as 
\begin{equation}\label{eq:Ew(i)}
\EE_{w}^{(i)}[\;\;\;] = \frac{\displaystyle
\int (\;\;\;)\;\prod_{j\neq i }^{n}p(X_{j}|w)^{\beta}\;\varphi(w)dw
}{\displaystyle
\int \prod_{j\neq i }^{n}p(X_{j}|w)^{\beta}\;\varphi(w)dw
},
\end{equation}
where $\displaystyle \prod_{j\neq i}^{n}$ 
shows the product for $j=1,2,3,.., n$, which does not include 
$j=i$. The predictive distribution 
leaving out $X_{i}$ is defined as 
\begin{equation}
p^{(i)}(x)=\EE_{w}^{(i)}[p(x|w)]. 
\end{equation}
The log loss of $p^{(i)}(x)$ when $X_{i}$ is used as a testing sample is
\begin{equation}
-\log p^{(i)}(X_{i})=-\log \EE_{w}^{(i)}[p(X_{i}|w)]. 
\end{equation}
Thus, the log loss of the Bayes cross-validation 
is defined as the empirical average of them, 
\begin{equation}\label{eq:Clv}
C_{v}L(n)=-\frac{1}{n}\sum_{i=1}^{n}
\log \EE_{w}^{(i)}[p(X_{i}|w)]. 
\end{equation}
The random variable $C_{v}L(n)$ is referred to as the {\it cross-validation loss}. 
Since $X_{1},X_{2},...,X_{n}$ are independent training samples, 
it immediately follows that
\begin{equation}
\EE[C_{v}L(n)]= \EE[B_{g}L(n-1)].
\end{equation}
Although the two random variables  $C_{v}L(n)$ and $B_{g}L(n-1)$ 
are different, 
\begin{equation}
C_{v}L(n)\neq B_{g}L(n-1),
\end{equation}
their expectation values coincide with each other by the definition. Using eq. (\ref{eq:BLWAIC}), it follows that
\begin{equation}
\EE[C_{v}L(n)]=\EE[\mbox{WAIC}(n-1)]+o(\frac{1}{n}).
\end{equation}
Therefore, three expectation values $\EE[C_{v}L(n)]$, $\EE[B_{g}L(n-1)]$, and $\EE[\mbox{WAIC}(n-1)]$ 
are asymptotically equal to each other. The primary goal of the present paper is to clarify
the asymptotic behaviors of 
three random variables, $C_{v}L(n)$, $B_{g}L(n)$, and $\mbox{WAIC}(n)$, when
$n$ is sufficiently large. 
\vskip3mm\noindent
{\bf Remark}. In practical applications, the Bayes generalization loss $B_{g}L(n)$ 
indicates the accuracy of Bayes estimation. However, in order to calculate 
$B_{g}L(n)$, we need the expectation value over the 
testing sample taken from the unknown true distribution, hence 
we cannot directly obtain $B_{g}L(n)$ in practical applications. 
On the other hand, both the cross-validation loss $C_{v}L(n)$ and 
the widely applicable information criterion $\mbox{WAIC}(n)$ can be calculated 
using only training samples. Therefore, the cross-validation loss and the
widely applicable information criterion can be used for model selection and hyperparameter
optimization. This is the reason why comparison of these random variables 
is an important problem in statistical learning theory. 

\section{Main Results} 

In this section, the main results of the present paper are explained. 
First, we define functional cumulants and describe their asymptotic properties. 
Second, we prove that both the cross-validation loss and the widely applicable information
criterion can be represented by the functional cumulants. 
Finally, we prove that the cross-validation loss and the widely applicable information
criterion are related to the birational invariants. 

\subsection{Functional Cumulants}

\noindent{\bf Definition}. The generating function $F(\alpha)$ 
of functional cumulants is defined as 
\begin{equation}
F(\alpha)=
\frac{1}{n}\sum_{i=1}^{n}\log \EE_{w}[p(X_{i}|w)^{\alpha}].
\end{equation}
The $k$th order functional cumulant $Y_{k}(n)$ $(k=1,2,3,4)$ is defined as 
\begin{equation}\label{eq:Yk}
Y_{k}(n)=\frac{d^{k}F}{d\alpha^{k}}(0).
\end{equation}
\vskip10mm
Then, by definition,
\begin{eqnarray}
F(0)&=&0,\\
F(1)&=&-B_{t}L(n),\\
Y_{1}(n)&=&-G_{t}L(n),\\
Y_{2}(n)&=&V(n)/n. 
\end{eqnarray}
For simple notation, we use 
\begin{equation}
\ell_{k}(X_{i})=\EE_{w}[(\log p(X_{i}|w))^{k}]\;\;\;(k=1,2,3,4). 
\end{equation}
\begin{Lemma}
Then, the following hold: 
\begin{eqnarray}
Y_{1}(n) & = & \frac{1}{n}\sum_{i=1}^{n}\ell_{1}(X_{i}),\label{eq:Y1} \\
Y_{2}(n) & = &  \frac{1}{n}\sum_{i=1}^{n}
\Bigl\{
\ell_{2}(X_{i})-\ell_{1}(X_{i})^{2}
\Bigr\},
\label{eq:Y2}\\
Y_{3}(n)& = & \frac{1}{n}\sum_{i=1}^{n}
\Bigl\{
\ell_{3}(X_{i})
-3\ell_{2}(X_{i})\ell_{1}(X_{i})
+2\ell_{1}(X_{i})^{3}\Bigr\},
\label{eq:Y3}\\
Y_{4}(n)& = & \frac{1}{n}\sum_{i=1}^{n}
\Bigl\{
\ell_{4}(X_{i})
-4\ell_{3}(X_{i})\ell_{1}(X_{i})
-3\ell_{2}(X_{i})^{2} \nonumber \\
&& +12\ell_{2}(X_{i})\ell_{1}(X_{i})^{2}
-6\ell_{1}(X_{i})^{4}
\Bigr\}. 
\label{eq:Y4}
\end{eqnarray}
Moreover, 
\begin{equation}
Y_{k}(n)=O_{p}(\frac{1}{n^{k/2}})\;\;\;(k=2,3,4).
\end{equation}
In other words, 
\begin{equation}\label{eq:Op}
{\rm limsup}_{n\rightarrow\infty}\EE[n^{k/2}\;|Y_{k}(n)|]<\infty\;\;\;
(k=2,3,4). 
\end{equation}
\end{Lemma}
\noindent(Proof) First, we prove 
Eqs. (\ref{eq:Y1}) through (\ref{eq:Y4}). 
Let us define
\begin{equation}
g_{i}(\alpha)=\EE_{w}[p(X_{i}|w)^{\alpha}]. 
\end{equation}
Then, $g_{i}(0)=1$, 
\begin{equation}
g_{i}^{(k)}(0)\equiv
\frac{d^{k}g_{i}}{d\alpha^{k}}(0)=\ell_{k}(X_{i})\;\;\;(k=1,2,3,4),
\end{equation}
and 
\begin{equation}
F(\alpha)=\frac{1}{n}\sum_{i=1}^{n}\log g_{i}(\alpha). 
\end{equation}
For arbitrary natural number $k$,
\begin{equation}
\Bigl(
\frac{g_{i}(\alpha)^{(k)}}{g_{i}(\alpha)}
\Bigr)'=
\frac{g_{i}(\alpha)^{(k+1)}}{g_{i}(\alpha)}
- 
\Bigl(
\frac{g_{i}(\alpha)^{(k)}}{g_{i}(\alpha)}
\Bigr)\Bigl(
\frac{g_{i}(\alpha)'}{g_{i}(\alpha)}\Bigr). 
\end{equation}
By applying this relation recursively, eqs.(\ref{eq:Y1}), 
(\ref{eq:Y2}), (\ref{eq:Y3}), and (\ref{eq:Y4}) are derived. 
Let us prove eq.(\ref{eq:Op}). The random variables $Y_{k}(n)$ $(k=2,3,4)$ are 
invariant under the transform,
\begin{equation}\label{eq:transf}
\log p(X_{i}|w)\mapsto 
\log p(X_{i}|w)+c(X_{i}), 
\end{equation}
for arbitrary $c(X_{i})$. In fact, by replacing $p(X_{i}|w)$ by
$p(X_{i}|w)e^{C(X_{i})}$, we define 
\begin{equation}
\hat{F}(\alpha)=\frac{1}{n}\sum_{i=1}^{n}
\log\EE_{w}[p(X_{i}|w)^{\alpha}\;e^{\alpha c(X_{i})}].
\end{equation}
Then, the difference between $F(\alpha)$ and $\hat{F}(\alpha)$ is a linear function
of $\alpha$, which vanishes by higher-order differentiation. 
In particular, by selecting $c(X_{i})=-\log p_{0}(X_{i})$, we can show that 
$Y_{k}(n)$ $(k=2,3,4)$ are invariant by the following replacement,
\begin{equation}
\log p(X_{i}|w)\mapsto 
f(X_{i},w). 
\end{equation}
In other words, $Y_{k}(n)$ $(n=2,3,4)$ are invarianrt by the replacement, 
\begin{equation}
\ell_{k}(X_{i})\mapsto \EE_{w}[f(X_{i},w)^{k}].
\end{equation}
Using the Cauchy-Schwarz inequality, for $1\leq k'\leq k$, 
\begin{equation}
\EE_{w}
[|f(X_{i},w)|^{k'}]^{1/k'}\leq \EE_{w}[|f(X_{i},w)|^{k}]^{1/k}.
\end{equation}
Therefore, for $k=2,3,4$, 
\begin{equation}
\EE[|Y_{k}(n)|]\leq 
\EE\Bigl[\frac{C_{k}}{n}
\sum_{i=1}^{n}\EE_{w}[|f(X_{i},w)|^{k}]
\Bigr]
\leq C_{k}M_{k}(n),
\end{equation}
where $C_{2}=2,C_{3}=6,C_{4}=26$. 
Then, using eq. (\ref{eq:assume}), we obtain eq. (\ref{eq:Op}). 
(Q.E.D.) 
\vskip10mm\noindent
{\bf Remark.} 
Using eq. (\ref{eq:transf}) with $c(X_{i})=-\EE_{w}[\log p(X_{i}|w)]$ and
the normalized function defined as
\begin{equation}
\ell_{k}^{*}(X_{i})=\EE_{w}[(\log p(X_{i}|w)-c(X_{i}))^{k}], 
\end{equation}
it follows that 
\begin{eqnarray}
Y_{2}(n) & = &  \frac{1}{n}\sum_{i=1}^{n}
\ell_{2}^{*}(X_{i}),\\
Y_{3}(n)& = & \frac{1}{n}\sum_{i=1}^{n}
\ell_{3}^{*}(X_{i}),
\\
Y_{4}(n)& = & \frac{1}{n}\sum_{i=1}^{n}
\Bigl\{
\ell_{4}^{*}(X_{i})
-3\ell_{2}^{*}(X_{i})^{2} 
\Bigr\}. 
\end{eqnarray}
These formulas may be useful in practical applications. 

\subsection{Bayes Cross-validation and Widely Applicable Information Criterion}

We show the asymptotic equivalence of the cross-validation loss 
$C_{v}L(n)$ and the widely applicable information criterion
$\mbox{WAIC}(n)$. 
\begin{Theorem}
For arbitrary $0<\beta<\infty$, the cross-validation loss 
$C_{v}L(n)$ and the widely applicable information 
criterion ${\rm{WAIC}}(n)$ are given, respectively, as
\begin{eqnarray}
C_{v}L(n)&=&-Y_{1}(n)+\Bigl(\frac{2\beta-1}{2}\Bigr)Y_{2}(n) \nonumber \\
&&-\Bigl(\frac{3\beta^{2}-3\beta+1}{6}\Bigr)Y_{3}(n) +O_{p}(\frac{1}{n^{2}}), \\
\mbox{\rm{WAIC}}(n)&=&-Y_{1}(n)+\Bigl(\frac{2\beta-1}{2}\Bigr)Y_{2}(n) \nonumber \\
&&-\frac{1}{6}Y_{3}(n) +O_{p}(\frac{1}{n^{2}}). 
\end{eqnarray}
\end{Theorem}
\noindent(Proof) First, we consider $C_{v}L(n)$. 
From the definitions of $\EE_{w}[\;\;]$ and $\EE_{w}^{(i)}[\;\;]$, we have
\begin{equation}\label{eq:post}
\EE_{w}^{(i)}[(\;\;\;)]=
\frac{\EE_{w}[(\;\;\;)p(X_{i}|w)^{-\beta}\;]
}{
\EE_{w}[p(X_{i}|w)^{-\beta}\;]
}.
\end{equation}
Therefore, by the definition of the cross-validation loss, eq. (\ref{eq:Clv}), 
\begin{equation}
C_{v}L(n)=-\frac{1}{n}\sum_{i=1}^{n}
\log \frac
{
\EE_{w}[\;p(X_{i}|w)^{1-\beta}\;]
}
{ \EE_{w} 
[\;p(X_{i}|w)^{-\beta}\;]
}
.
\end{equation}
Using the generating function of functional cumulants $F(\alpha)$, 
\begin{equation}
C_{v}L(n)=F(-\beta)-F(1-\beta).\label{eq:sasasa}
\end{equation}
Then, using Lemma \ref{Lemma:1} (2) for each $k=2,3,4$, 
and $|\alpha|<1+\beta$, 
\begin{eqnarray}
\EE[|F^{(k)}(\alpha)|] & \leq & 
\EE\Bigl[
\frac{C_{k}}{n}
\sum_{i=1}^{n}
\frac{\EE_{w}[|f(X_{i},w)|^{k}\exp(\alpha f(X_{i},w))]}
{\EE_{w}[\exp(\alpha f(X_{i},w))]} 
\Bigr]
\nonumber \\
& \leq & C_{k}M_{k}(n),
\end{eqnarray}
where $C_{2}=2,C_{3}=6,C_{4}=26$. 
Therefore, 
\begin{equation}\label{eq:Fk}
|F^{(k)}(\alpha)|=O_{p}(\frac{1}{n^{k/2}}). 
\end{equation}
By Taylor expansion of $F(\alpha)$ among $\alpha=0$, 
there exist $\beta^{*},\beta^{**}$ ($|\beta^{*}|,|\beta^{**}|<1+\beta$) 
such that 
\begin{eqnarray}
F(-\beta)&=& F(0) -\beta F'(0)
+\frac{\beta^{2}}{2}F''(0) \nonumber \\
&& -\frac{\beta^{3}}{6}F^{(3)}(0)
+\frac{\beta^{4}}{24}F^{(4)}(\beta^{*}), \\
F(1-\beta)&=& F(0)+(1-\beta) F'(0)
+\frac{(1-\beta)^{2}}{2}F''(0) \nonumber \\
&&+\frac{(1-\beta)^{3}}{6}F^{(3)}(0)
+\frac{(1-\beta)^{4}}{24}F^{(4)}(\beta^{**}).
\end{eqnarray}
Using $F(0)=0$ and Eqs. (\ref{eq:sasasa}) and (\ref{eq:Fk}), it follows that 
\begin{eqnarray}
C_{v}L(n)
&=& -F'(0)
+\frac{2\beta-1}{2}F''(0)\nonumber \\
&& -\frac{3\beta^{2}-3\beta+1}{6}F^{(3)}(0) 
+O_{p}(\frac{1}{n^{2}}).
\end{eqnarray}
Thus, we have proven the first half of the theorem. 
For the latter half, by the definitions of $\mbox{WAIC}(n)$, 
Bayes training loss, and the functional variance, we have 
\begin{eqnarray}
\mbox{WAIC}(n) &=&  B_{t}L(n)+(\beta/n) V(n),\\
B_{t}L(n)&=&-F(1), \\
V(n)&=&n F''(0).
\end{eqnarray}
Therefore,
\begin{equation}
\mbox{WAIC}(n)=-F(1)+\beta  F''(0).
\end{equation}
By Taylor expansion of $F(1)$, we obtain
\begin{equation}
\mbox{WAIC}(n)= -F'(0)
+\frac{2\beta-1}{2}F''(0)
-\frac{1}{6}F^{(3)}(0) 
+O_{p}(\frac{1}{n^{2}}),
\end{equation}
which completes the proof. (Q.E.D.) 
\vskip5mm
From the above theorem, we obtain the following corollary. 
\begin{Corollary}\label{corollary:main}
For arbitrary $0<\beta<\infty$, the cross-validation loss 
$C_{v}L(n)$ and the widely applicable information criterion 
${\rm{WAIC}}(n)$ satisfy
\begin{equation}
C_{v}L(n)={\rm{WAIC}}(n)+O_{p}(\frac{1}{n^{3/2}}). 
\end{equation}
In particular, for $\beta=1$, 
\begin{equation}
C_{v}L(n)={\rm{WAIC}}(n)+O_{p}(\frac{1}{n^{2}}). 
\end{equation}
\end{Corollary}
More precisely, the difference between the cross-validation loss 
and the widely applicable information criterion is given by 
\begin{equation}
C_{v}L(n)-{\rm{WAIC}}(n)
\cong
\Bigl(\frac{\beta-\beta^{2}}{2}\Bigr)Y_{3}(n). 
\end{equation}
If $\beta=1$, 
\begin{equation}
C_{v}L(n)-{\rm{WAIC}}(n)
\cong
\frac{1}{12}Y_{4}(n). 
\end{equation}

\subsection{Generalization Error and Cross-validation Error}

In the previous subsection, we have shown that
the cross-validation loss is asymptotically
equivalent to the widely applicable information criterion.
In this section, let us compare the Bayes generalization error $B_{g}(n)$ 
given in eq. (\ref{eq:Bg}) 
and the cross-validation error $C_{v}(n)$, which is defined as 
\begin{equation}
C_{v}(n)=C_{v}L(n)-L_{n}.\label{eq:Cv}
\end{equation}
We need mathematical concepts, the real log canonical threshold,
and the singular fluctuation.  
\vskip3mm\noindent
{\bf Definition}. The zeta function $\zeta(z)$ $(Re(z)>0)$ 
of statistical learning is 
defined as 
\begin{equation}
\zeta(z)=\int K(w)^{z}\varphi(w)dw,
\end{equation}
where
\begin{equation}
K(w)=\EE_{X}[f(X,w)]
\end{equation}
is a nonnegative analytic function. Here,
$\zeta(z)$ can be analytically continued to 
the unique meromorphic function on the entire complex plane ${\CC}$. 
All poles of $\zeta(z)$ are real, negative, and rational numbers.
The maximum pole is denoted as
\begin{equation}\label{eq:lambda}
(-\lambda)=\mbox{ maximum pole of } \zeta(z). 
\end{equation}
Then, the positive rational number $\lambda$ 
is referred to as the {\it real log canonical threshold}. The 
{\it singular fluctuation} is defined as
\begin{equation}\label{eq:nyu}
\nu=\nu(\beta)=\lim_{n\rightarrow\infty}
\frac{\beta}{2} \EE[V(n)]. 
\end{equation}
Note that the real log canonical threshold does not depend on $\beta$, whereas 
the singular fluctuation is a function of $\beta$. 
\vskip10mm
Both the real log canonical threshold and the singular fluctuation 
are birational invariants. In other words, they are determined by the 
algebraic geometrical structure of the statistical model. 
The following lemma was proven in a previous study \cite{NN2010,IEICE2010,IWSMI2010}. 
\begin{Lemma}\label{Lemma:bgbt}
The following convergences hold: 
\begin{eqnarray}
\lim_{n\rightarrow\infty} n\EE[B_{g}(n)]&=&\frac{\lambda-\nu}{\beta}+\nu, \label{eq:ln1}\\
\lim_{n\rightarrow\infty} n\EE[B_{t}(n)]&=&\frac{\lambda-\nu}{\beta}-\nu, \label{eq:ln2}
\end{eqnarray}
Moreover, convergence in probability
\begin{equation}\label{eq:BGBTV}
n(B_{g}(n)+B_{t}(n))+V(n)\rightarrow\frac{2\lambda}{\beta}
\end{equation}
holds.
\end{Lemma}
\noindent(Proof) 
For the case in which $q(x)$ is realizable by and singular for $p(x|w)$, 
eqs. (\ref{eq:ln1}) and (\ref{eq:ln2}) were proven by 
in Corollary 3 in \cite{NN2010}. 
The equation (\ref{eq:BGBTV}) was given in Corollary 2 in \cite{NN2010}. 
For the case in which $q(x)$ is regular for $p(x|w)$, 
these results were
proved in \cite{IEICE2010}. For the case in which $q(x)$ is
singular for and unrealizable by $p(x|w)$ they were generalized in \cite{IWSMI2010}. 
(Q.E.D.) 
\vskip5mm\noindent
{\bf Examples}. If $q(x)$ is regular for and realizable by $p(x|w)$, then
$\lambda=\nu=d/2$, where $d$ is the dimension of the parameter space.
If $q(x)$ is regular for and unrealizable by $p(x|w)$, then 
$\lambda$ and $\nu$ are given by \cite{IEICE2010}. 
If $q(x)$ is singular for and realizable by $p(x|w)$, then 
$\lambda$ for several models are obtained by resolution of singularities 
\cite{Aoyagi,Rusakov,Yamazaki,Lin,Zwiernik}.
If $q(x)$ is singular for and unrealizable by $p(x|w)$, then $\lambda$ and $\nu$ remain unknown constants. 
\vskip3mm
We have the following theorem. 
\begin{Theorem}\label{Theorem:222}
The following equation holds:
\begin{equation}
\lim_{n\rightarrow\infty} n\EE[C_{v}(n)]=\frac{\lambda-\nu}{\beta}+\nu, \label{eq:ln3}
\end{equation}
The sum of the Bayes generalization error and the cross-validation error
satisfies 
\begin{equation}
B_{g}(n)+C_{v}(n)
=(\beta-1)\frac{V(n)}{n}+\frac{2\lambda}{\beta n}+o_{p}(\frac{1}{n}).  
\end{equation}
In particular, if $\beta=1$,
\begin{equation}
B_{g}(n)+C_{v}(n)
=\frac{2\lambda}{n}+o_{p}(\frac{1}{n}). 
\end{equation}
\end{Theorem}
\noindent(Proof) By eq. (\ref{eq:ln1}), 
\begin{equation}
\EE[B_{g}(n-1)]=
\Bigl(\frac{\lambda-\nu}{\beta}+\nu
\Bigr)\frac{1}{n}+o(\frac{1}{n}). 
\end{equation}
Since $\EE[C_{v}(n)]=\EE[B_{g}(n-1)]$, 
\begin{eqnarray}
\lim_{n\rightarrow\infty} n\EE[C_{v}(n)]&=&
\lim_{n\rightarrow\infty} n\EE[B_{g}(n-1)]\\
&=& 
\frac{\lambda-\nu}{\beta}+\nu.
\end{eqnarray} 
From eq. (\ref{eq:BGBTV}) and 
Corollary \ref{corollary:main}, 
\begin{equation}
B_{t}(n)=C_{v}(n)-\frac{\beta}{n} V(n)+O_{p}(\frac{1}{n^{3/2}}),
\end{equation}
and it follows that 
\begin{equation}
(B_{g}(n)+C_{v}(n))
=(\beta-1)\frac{V(n)}{n}+\frac{2\lambda}{\beta n}+o_{p}(\frac{1}{n}),
\end{equation}
which proves the Theorem. 
(Q.E.D.) 
\vskip5mm
This theorem indicates that both the cross-validation error and the
Bayes generalization error are determined by the 
algebraic geometrical structure of the statistical model,
which is extracted as the real log canonical threshold. 
From this theorem, in the strict Bayes case $\beta=1$, we have
\begin{eqnarray}
\EE[B_{g}(n)] & = &\frac{\lambda}{n}+o(\frac{1}{n}),\\
\EE[C_{v}(n)] & = & \frac{\lambda}{n}+o(\frac{1}{n}),
\end{eqnarray}
and
\begin{equation}\label{eq:bgcv}
B_{g}(n)+C_{v}(n)=\frac{2\lambda}{n}+o_{p}(\frac{1}{n}).
\end{equation}
Therefore, the smaller cross-validation error $C_{v}(n)$ is equivalent to the 
larger Bayes generalization error $B_{g}(n)$. Note that a regular statistical model
is a special example of singular models, hence both Theorems 1 and 2 also hold 
in regular statistical models. 
In \cite{Cambridge2009}, it was proven that the random variable $nB_{g}(n)$ 
converges to a random variable in law. Thus, $nC_{v}(n)$ converges to 
a random variable in law. The asymptotic probability distribution of 
$nB_{g}(n)$ can be represented using a Gaussian process, which is defined on
the set of true parameters, but is 
not equal to the $\chi^{2}$ distribution in general. 
\vskip3mm\noindent
{\bf Remark.} The relation given by eq. (\ref{eq:bgcv}) indicates that, if $\beta=1$, 
the variances of $B_{g}(n)$ and $C_{v}(n)$ 
are equal. 
If the average value $2\nu=\EE[ V(n)]$ is known, then 
$B_{t}(n)+2\nu/n$ can be used instead of $C_{v}(n)$, because 
both average values are asymptotically equal to the Bayes generalization error. 
The variance of $B_{t}(n)+2\nu/n$ is smaller than that of $C_{v}(n)$ 
if and only if the variance of $B_{t}(n)$ is smaller than that of $B_{g}(n)$.
If a true distribution is regular for and realizable by the statistical
model, then the variance of $B_{t}(n)$ is asymptotically equal to that of $B_{g}(n)$. 
However, in other cases, the variance of $B_{t}(n)$ may be smaller or larger than 
that of $B_{g}(n)$. 

\section{Discussion}

\label{section:discuss}

Let us now discuss the results of the present paper.

\subsection{From Regular to Singular}

First, we summarize the regular and singular learning theories. 

In regular statistical models, 
the generalization loss of the maximum likelihood method 
is asymptotically equal to that of the Bayes estimation. 
In both the maximum likelihood and Bayes methods, 
the cross-validation losses have the same asymptotic behaviors. 
The leave-one-out cross-validation is
asymptotically equivalent to the AIC, 
in both the maximum likelihood and Bayes methods.  

On the other hand, in singular learning machines, 
the generalization loss of the maximum likelihood method
is larger than the Bayes generalization loss. 
Since the generalization loss of the maximum likelihood method 
is determined by the maximum value of the Gaussian process,
the maximum likelihood method is not 
appropriate in singular models \cite{Cambridge2009}. 
In Bayes estimation, we derived the asymptotic expansion 
of the generalization loss and proved that the average of 
the widely applicable information criterion is asymptotically
equal to the Bayes generalization loss \cite{NN2010}. 
In the present paper, we clarified that the leave-one-out cross-validation 
in Bayes estimation is asymptotically equivalent to WAIC. 

It was proven \cite{NC2001} that the Bayes marginal likelihood
of a singular model is different from BIC of a regular model.
In the future, we intend to compare the cross-validation and
Bayes marginal likelihood in model selection and hyperparameter 
optimization in singular statistical models. 

\subsection{Cross- validation and Importance Sampling}

Second, let us investigate the cross-validation and the importance sampling 
cross-validation from a practical viewpoint. 

In Theorem 1, we theoretically proved that the leave-one-out cross-validation 
is asymptotically equivalent to the widely applicable 
information criterion. In practical applications, 
we often approximate the posterior 
distribution using the Markov Chain Monte Carlo or other 
numerical methods. If the posterior distribution is 
precisely realized, then the two theorems of the present paper hold. 
However, if the posterior distribution was not precisely
approximated, then the cross-validation might not be equivalent to 
the widely applicable information criterion. 

In Bayes estimation, there are two different methods by which 
the leave-one-out cross-validation is numerically approximated.  
In the former method, $CV_{1}$ is obtained by realizing all posterior 
distributions $\EE_{w}^{(i)}[\;\;]$ leaving out $X_{i}$ 
for $i=1,2,3,...,n$, and the empirical average
\begin{equation}
CV_{1}=-\frac{1}{n}\sum_{i=1}^{n}\log \EE_{w}^{(i)}[p(X_{i}|w)]
\end{equation}
is then calculated. In this method, we must realize $n$ different 
posterior distributions, which requires heavy computational costs. 

In the latter method, the posterior distribution leaving out $X_{i}$
is estimated using the posterior average 
$\EE_{w}[\;\;]$, in the same manner as eq. (\ref{eq:post}), 
\begin{equation}
\EE_{w}^{(i)}[p(X_{i}|w)]\cong 
\frac{\EE_{w}[p(X_{i}|w)\;p(X_{i}|w)^{-\beta}\;]
}{
\EE_{w}[p(X_{i}|w)^{-\beta}\;]
}.
\end{equation}
This method is referred to as the importance sampling leave-one-out cross-validation \cite{Gelfand2}, in which only one posterior distribution is needed and 
the leave-one-out cross-validation is approximated by $CV_{2}$, 
\begin{equation}
CV_{2}\cong -\frac{1}{n}\sum_{i=1}^{n}\log \frac{\EE_{w}[p(X_{i}|w) \;p(X_{i}|w)^{-\beta}\;]
}{
\EE_{w}[p(X_{i}|w)^{-\beta}\;]
}.
\end{equation}

If the posterior distribution is completely realized, 
then $CV_{1}$ and $CV_{2}$ coincide with each other and are asymptotically
equivalent to the widely applicable information criterion. 
However, if the posterior distribution is not sufficiently approximated,
then the values $CV_{1}$, $CV_{2}$, and $\mbox{WAIC}(n)$ might be different. 

The average values using the posterior distribution may sometimes have
infinite variances \cite{Peruggia} if the set of parameters is not compact. 
Moreover, in singular learning machines, the set of true parameters 
is not a single point but rather an analytic set, hence we must restrict 
the parameter space to be compact for well-defined average values. 
Therefore, we adopted the assumptions in Subsection \ref{subsection:assumption}
that the parameter space is compact and the log likelihood function has 
the appropriate properties. Under these conditions, 
the observables studied in the present paper have finite variances.

\subsection{Comparison with the Deviance Information Criteria}

Third, let us compare the deviance information criterion (DIC) 
\cite{DIC} to the Bayes cross-validation
and WAIC, because DIC 
is sometimes used in Bayesian model evaluation. 
In order to estimate the Bayesian generalization error, 
DIC is written by
\begin{equation}
DIC_{1}=B_{t}L(n)+\frac{2}{n}\sum_{i=1}^{n}
\Bigl{\{}
-E_{w}[\log p(X_{i}|w)]+\log p(X_{i}|E_{w}[w])
\Bigr{\}},
\end{equation}
where the second term of the right-hand side corresponds to 
the ``effective number of parameters" of
DIC divided by the number of parameters. 
Under the condition that the log likelihood ratio function in the posterior distribution 
is subject to the $\chi^{2}$ distribution, a modified DIC 
was proposed \cite{Gelman} as
\begin{equation}
DIC_{2}=B_{t}L(n)+
\frac{2}{n}
\Bigl{[}
E_{w}\bigl{[}\Bigl{\{}\sum_{i=1}^{n}\log p(X_{i}|w)\Bigr{\}}^{2}
\bigr{]}
-E_{w}\bigl{[}\sum_{i=1}^{n}\log p(X_{i}|w)\bigr{]}^{2}
\Bigr{]},
\end{equation}
the variance of which was investigated previously \cite{Raftery}. 
Note that $DIC_{2}$ is different from $\mbox{WAIC}$. In a singular
learning machine, since the set of optimal parameters 
is an analytic set, the correlation between different true parameters does not vanish,
even asymptotically. 

%%%%((lambda-nu)/beta-nu + 2(lambda/beta-nu) + 2nu

We first derive the theoretical properties of DIC. 
If the true distribution is regular for the statistical model, 
then the set of the optimal parameter is a single point $w_{0}$. Thus, the difference
of $E_{w}[w]$ and the maximum {\it a posteriori} 
estimator is asymptotically smaller than $1/\sqrt{n}$. Therefore, 
based on the results in \cite{IEICE2010}, if $\beta=1$, 
\begin{equation}
\EE[DIC_{1}]=L_{0}+(
3\lambda-2\nu(1)) \frac{1}{n}+o(\frac{1}{n}).
\end{equation}
If the true distribution is realizable by or regular for the statistical
model and if $\beta=1$, then the asymptotic behavior of $DIC_{2}$ is given by 
\begin{equation}\label{eq:DIC2-asymp}
\EE[DIC_{2}]=L_{0}+(
3\lambda-2\nu(1)+2\nu'(1)
)\frac{1}{n}+o(\frac{1}{n}),
\end{equation}
where $\nu'(1)=(d\nu/d\beta)(1)$. Equation (\ref{eq:DIC2-asymp}) 
is derived from the relations \cite{Cambridge2009,NN2010,IEICE2010,IWSMI2010},
\begin{eqnarray}
DIC_{2}& =& B_{t}L(n)-2\frac{\partial}{\partial\beta}G_{t}L(n), \\
\EE[G_{t}L(n)]&=&L_{0}+\Bigl{(}\frac{\lambda}{\beta}-\nu(\beta)\Bigr{)}\frac{1}{n} 
+o(\frac{1}{n}), 
\end{eqnarray}
where $G_{t}L(n)$ is given by eq. (\ref{eq:Gibbs}). 

Next, let us consider the DIC for each case. 
If the true distribution is regular for and realizable by the 
statistical model and if $\beta=1$, then $\lambda=\nu=d/2$, $\nu'(1)=0$, 
where $d$ is the number of parameters. Thus, 
their averages are asymptotically equal to the Bayes generalization error, 
\begin{eqnarray}
\EE[DIC_{1}]&=&L_{0}+\frac{d}{2n}+o(\frac{1}{n}), \\
\EE[DIC_{2}]&=&L_{0}+\frac{d}{2n}+o(\frac{1}{n}).
\end{eqnarray}
In this case, the averages of 
$DIC_{1}$, $DIC_{2}$,
$CV_{1}$, $CV_{2}$, and $\mbox{WAIC}$ have the same asymptotic behavior.

If the true distribution is regular for and unrealizable by
the statistical model and if $\beta=1$, then 
$\lambda=d/2$, $\nu= \mbox{tr}(IJ^{-1})$, 
and $\nu'(1)=0$ \cite{IEICE2010}, 
where $I$ is the Fisher information
matrix at $w_{0}$, and $J$ is the Hessian matrix of $L(w)$ at $w=w_{0}$. 
Thus, we have
\begin{eqnarray}
\EE[DIC_{1}]&=&L_{0}+\Bigl(\frac{3d}{2}-\mbox{tr}(IJ^{-1})\Bigr)\frac{1}{n}+o(\frac{1}{n}), \\
\EE[DIC_{2}]&=&L_{0}+\Bigl(\frac{3d}{2}-\mbox{tr}(IJ^{-1})\Bigr)\frac{1}{n}+o(\frac{1}{n}).
\end{eqnarray}
In this case, as shown in Lemma \ref{Lemma:bgbt},
the Bayes generalization error 
is given by $L_{0}+d/(2n)$ asymptotically, 
and so the averages of the deviance information 
criteria are not equal to the average of the Bayes generalization error. 

If the true distribution is singular for and realizable by
the statistical model and if $\beta=1$, then 
\begin{eqnarray}
\EE[DIC_{1}]&=&C+o(1), \label{eq:CCC} \\
\EE[DIC_{2}]&=&L_{0}+(3\lambda-2\nu(1)+2\nu'(1))\frac{1}{n}+o(\frac{1}{n}), 
\label{eq:dic3}
\end{eqnarray}
where $C$ $(C\neq L_{0})$ is, in general, a constant. Equation (\ref{eq:CCC}) 
is obtained because the set of true parameters 
in a singular model is not a single point, but rather an analytic set, 
so that, in general, the average $E_{w}[w]$ is not contained in the
neighborhood of the set of the true parameters. 
Hence the averages of the  deviance information criteria are not equal to 
those of the Bayes generalization error. 

The averages of the cross-validation loss 
and WAIC have the same asymptotic behavior 
as that of the Bayes generalization error, even if the 
true distribution is unrealizable by or singular for the statistical model.
Therefore, the deviance information criteria are different from the cross-validation
and WAIC, if the true distribution is singular for or unrealizable by the statistical model.  

\subsection{Experiment}

In this section, we describe an experiment. 
The purpose of the present paper is to clarify the theoretical properties of
the cross-validation and the widely applicable information criterion. 
An experiment was conducted in order to illustrate the main theorems. 

Let $x,y\in {\RR}^{3}$. We considered a statistical model defined as 
\begin{equation}
p(x,y|w)=\frac{s(x)}{(2\pi\sigma^{2})^{3/2}}
\exp(-\frac{\|y-R_{H}(x,w)\|^{2}}{2\sigma^{2}}),
\end{equation}
where $\sigma=0.1$ and $s(x)$ is ${\cal N}(0,2^{2}I)$.
Here, ${\cal N}(m,A)$ exhibits a 
normal distribution with the average vector $m$ and the covariance 
matrix $A$, and $I$ is the identity matrix. 
Note that the distribution $s(x)$ was not estimated. 
We used a three-layered neural network, 
\begin{equation}
R_{H}(x,w)=\sum_{h=1}^{H}a_{h}\tanh(b_{h}\cdot x),
\end{equation}
where the parameter was 
\begin{equation}
w=\{(a_{h}\in {\RR}^{3},b_{h}\in{\RR}^{3})\;;\;h=1,2,...,H\}
\in {\RR}^{6H}.
\end{equation}
In the experiment, a learning machine with $H=3$ was used and
the true distribution was set with $H=1$. The parameter that 
gives the distribution is denoted as $w_{0}$, which denotes 
the parameters of both models $H=1,3$. 
Then, $R_{H}(x,w_{0})=R_{H_{0}}(x,w_{0})$. Under this condition, 
the set of true parameters
\begin{equation}
\{w\in W;p(x|w)=p(x|w_{0})\}
\end{equation}
is not a single point but an analytic set with singularities, 
resulting that the regularity condition is not satisfied. 
In this case, the log density ratio function is equivalent to 
\begin{equation}
f(x,y,w)=\frac{1}{2\sigma^{2}}
\Bigr{\{} 
\|y-R_{H}(x,w)\|^{2}-\|y-R_{H}(x,w_{0})\|^{2}
\Bigl{\}}.
\end{equation}
In this model, although the Bayes generalization error is not 
equal to the average square error
\begin{equation}
\mbox{SE}(n)=\frac{1}{2\sigma^{2}}\EE \EE_{X}
\Bigl[\|\;R_{H}(X,w_{0})-\EE_{w}[R_{H}(X,w)]\;\|^{2}\Bigr],
\end{equation}
asymptotically $\mbox{SE}(n)$ and $B_{g}(n)$ are equal to each other 
\cite{Cambridge2009}. 

The prior distribution 
$\varphi(w)$ was set as ${\cal N}(0,10^{2}I)$. Although this 
prior does not have compact support mathematically, it can be
understood in the experiment that the support of $\varphi(w)$ 
is essentially contained in a sufficiently large compact set. 

In the experiment, the number of training samples was fixed as $n=200$.
One hundred sets of 200 training samples each were obtained independently. 
For each training set, the strict Bayes posterior distribution $\beta=1$ was
approximated by the Markov chain Monte Carlo (MCMC) method.
The Metropolis method, in which each random trial was 
taken from ${\cal N}(0,(0.005)^{2}I)$, was applied, and 
the average exchanging ratio was obtained as approximately 0.35. 
After 100,000 iterations of Metropolis random sampling, 200 parameters were obtained in every 100 sampling steps. For a fixed training set, by changing the initial values
and the random seeds of the software, the same MCMC sampling procedures were performed 10 
times independently, which was done 
for the purpose of minimizing the effect of the local minima. 
Finally, for each training set,
we obtained $200\times10=2,000$ parameters, which were used to
approximate the posterior distribution. 

Table 2 shows the experimental results. We observed 
the Bayes generalization error $BG=B_{g}(n)$, 
the Bayes training error $BT=B_{t}(n)$,
importance sampling leave-one-out cross-validation $CV=CV_{2}-L_{n}$, 
the widely applicable information criterion $\mbox{WAIC}=\mbox{WAIC}(n)-L_{n}$, 
two deviance information criteria, namely, $DIC1=DIC_{1}-L_{n}$ and $DIC2=DIC_{2}-L_{n}$,
 and the sum $BG+CV=B_{g}(n)+C_{v}(n)$. The values $AVR$ and $STD$ in Table 2 
show the average and standard deviation of one hundred sets of training data, respectively. 
The original cross-validation $CV_{1}$ was not 
observed because the associated computational cost was too high. 

The experimental results reveal that the average and standard deviation of
$BG$ were approximately the same as those of $CV$ and $\mbox{WAIC}$, which 
indicates that Theorem 1 holds. The real log canonical threshold, 
the singular fluctuation, and its derivative of this case were estimated as
\begin{eqnarray}
\lambda&\approx&5.6, \\
\nu(1) &\approx &7.9, \\
\nu'(1)&\approx & 3.6.
\end{eqnarray}
Note that, if the true distribution is regular for and realizable by
the statistical model, $\lambda=\nu(1)=d/2=9$ and $\nu'(1)=0$.
The averages of the two deviance information criteria were not equal to 
that of the Bayes generalization error. 
The standard deviation of $BG+CV$ was smaller than 
the standard deviations of $BG$ and $CV$, which is in agreement with Theorem \ref{Theorem:222}. 

Note that the standard deviation of $BT$ was 
larger than those of $CV$ and $\mbox{WAIC}$, which indicates that, 
even if the average value $\EE[C_{v}(n)-B_{t}(n)]=2\nu/n$ is known 
and an alternative cross-validation, such as the AIC,
\begin{equation}
CV_{3}=B_{t}L(n)+2\nu/n,
\end{equation}
is used, then the variance of $CV_{3}-L_{n}$ was larger than 
the variances of $C_{v}L(n)-L_{n}$ and $\mbox{WAIC}(n)-L_{n}$. 

\begin{table}[tb]\label{table:222}
\begin{center}
\begin{tabular}{|c|c|c|c|c|c|c|c|}
\hline
    & $BG$ & $BT$ & $CV$ & $\mbox{WAIC}$ & $DIC1$ & $DIC2$ & $BG+CV$ \\
\hline
AVR & 0.0264  & -0.0511& 0.0298 & 0.0278 & -35.1077 & 0.0415 & 0.0562  \\
\hline
STD & 0.0120  & 0.0165 & 0.0137 & 0.0134 &  19.1350 & 0.0235 & 0.0071  \\
\hline
\end{tabular}
\end{center}
\caption{Average and standard deviation}
\end{table}

Table 3 shows the correlation matrix for several values. 
The correlation between $CV$ and $\mbox{WAIC}$ was 0.996, which indicates that
Theorem 1 holds. The correlation between $BG$ and $CV$ was -0.854, 
and that between $BG$ and $\mbox{WAIC}$ was -0.873, which corresponds to Theorem 2.

\begin{table}[tb]\label{table:333}
\begin{center}
\begin{tabular}{|c|c|c|c|c|c|c|c|}
\hline
    & $BG$ & $BT$ & $CV$ & $\mbox{WAIC}$ & $DIC1$ & $DIC2$ & $BG+CV$ \\
\hline
$BG$& 1.000 & -0.854 &  -0.854 & -0.873 & 0.031 & -0.327 & 0.043\\
\hline 
$BT$& &  1.000 & 0.717 &  0.736 & 0.066 & 0.203 & -0.060 \\
\hline
$CV$ & & & 1.000 &  0.996 & -0.087 & 0.340 & 0.481 \\
\hline
$WA$ &  &  &  & 1.000 & -0.085 & 0.341 & 0.443\\
\hline
$DIC1$ & &  &  &  & 1.000& -0.069 & -0.115 \\
\hline
$DIC2$ &  & &  & &  & 1.000& 0.102 \\
\hline
\end{tabular}
\end{center}
\caption{Correlation matrix}
\end{table}

The accuracy of numerical approximation of
the posterior distribution depends on 
the statistical model, the true distribution, the prior distribution, 
the Markov chain Monte Carlo method, and the experimental fluctuation. 
In the future, we intend to develop a method by which to design experiments. 
The theorems proven in the present paper may be useful in such research. 

\subsection{Birational Invariant}

Finally, we investigate the statistical problem from an algebraic geometrical viewpoint. 

In Bayes estimation, we can introduce an analytic function of the
parameter space $g:U\rightarrow W$, 
\begin{equation}
w=g(u).
\end{equation}
Let $|g'(u)|$ be its Jacobian determinant. Note that
the inverse function $g^{-1}$ is not needed if $g$ satisfies
the condition that 
$\{u\in U;|g'(u)|=0\}$ is a measure zero set in $U$. Such a 
function $g$ is referred to as a birational transform. 
It is important that, by the transform, 
\begin{eqnarray}
p(x|w) & \mapsto & p(x|g(u)), \\
\varphi(w)&\mapsto &\varphi(g(u))|g'(u)|,
\end{eqnarray}
the Bayes estimation on $W$ is equivalent to that on $U$. 
A constant defined for a set of statistical models and a prior is
said to be a birational invariant if it is invariant under such a transform 
$w=g(u)$. 

The real log canonical threshold $\lambda$ is a birational invariant 
\cite{Atiyah,Hironaka,Kashiwara,Kollor,Mustata,Cambridge2009} 
that represents the algebraic geometrical relation between the set of 
parameters $W$ and the set of the optimal parameters $W_{0}$. 
Although the singular fluctuation is also a birational invariant, 
its properties remain unknown. 
In the present paper, we proved in Theorem 1 that
\begin{equation}\label{eq:heikin}
\EE[B_{g}L(n)]=\EE[C_{v}L(n)]+o(1/n).
\end{equation}
On the other hand, in Theorem 2, we proved that
\begin{equation}\label{eq:randomv}
B_{g}(n)+C_{v}(n)=\frac{2\lambda}{n}+o_{p}(1/n). 
\end{equation}
In model selection or hyperparameter optimization, eq. (\ref{eq:heikin}) 
shows that minimization 
of the cross-validation makes the generalization loss smaller on average. 
However, eq. (\ref{eq:randomv}) shows that minimization of
the cross-validation does not ensure minimum generalization loss. 
The widely applicable information criterion has the same property as 
the cross-validation. The constant $\lambda$ appears to exhibit a bound, 
which can be attained by statistical estimation for a given pair of
a statistical model and a prior distribution. Hence, 
clarification of the algebraic geometrical structure in statistical estimation 
is an important problem in statistical learning theory. 

\section{Conclusion}

In the present paper, we have shown theoretically that the leave-one-out cross-validation 
in Bayes estimation 
is asymptotically equal to the widely applicable information criterion
and that the sum of the cross-validation error and the generalization error 
is equal to twice the real log canonical threshold divided by the number of 
training samples. In addition, we clarified that cross-validation and the widely applicable
information criterion are different from the deviance information criteria. 
This result indicates that, even in singular statistical
models, the cross-validation is asymptotically equivalent to the information 
criterion, and that the asymptotic properties of these models are determined by the 
algebraic geometrical structure of a statistical model.

\end{document}